\pdfoutput=1

\documentclass[11pt]{article}

\usepackage[]{acl}

\usepackage{times}
\usepackage{latexsym}
\usepackage{xcolor}
\usepackage[T1]{fontenc}

\usepackage[utf8]{inputenc}

\usepackage{microtype}

\usepackage{inconsolata}

\usepackage{graphicx}
\usepackage{amsmath}
\usepackage{cleveref}
\usepackage{twemojis}
\usepackage{booktabs}
\usepackage{multirow}
\usepackage{subcaption}
\usepackage{enumitem}
\usepackage{listings}
\title{GlotEval: A Test Suite for Massively Multilingual Evaluation of Large Language Models}
\author{
    \textbf{Hengyu Luo\textsuperscript{1}},  
    \textbf{Zihao Li\textsuperscript{1}},  
    \textbf{Joseph Attieh\textsuperscript{1$\dag$}},  
    \textbf{Sawal Devkota\textsuperscript{2$\dag$}},  
    \textbf{Ona de Gibert\textsuperscript{1$\dag$}}, \\
    \textbf{Xu Huang\textsuperscript{4$\dag$}}, 
    \textbf{Shaoxiong Ji\textsuperscript{2,5,6$\dag$}},  
    \textbf{Peiqin Lin\textsuperscript{3$\dag$}},  
    \textbf{Bhavani Sai Praneeth Varma Mantina\textsuperscript{2$\dag$}}, \\  
    \textbf{Ananda Sreenidhi\textsuperscript{2$\dag$}},
    \textbf{Raúl Vázquez\textsuperscript{1$\dag$}},
    \textbf{Mengjie Wang\textsuperscript{2$\dag$}},  
    \textbf{Samea Yusofi\textsuperscript{2$\dag$}}, \\
    \textbf{Fei Yuan\textsuperscript{7}},
    \textbf{J\"org Tiedemann\textsuperscript{1,5}}
    \\
    \textsuperscript{1}University of Helsinki, Finland  
    \textsuperscript{2}Technical University of Darmstadt, Germany \\  
    \textsuperscript{3}University of Munich, Germany  
    \textsuperscript{4}Nanjing University, China 
    \textsuperscript{5}ELLIS Institute Finland \\
    \textsuperscript{6}University of Turku, Finland 
    \textsuperscript{7}Shanghai Artificial Intelligence Laboratory, China
    \\
    \small{
        \textbf{Correspondence:} \href{mailto:hengyu.luo@helsinki.fi}{hengyu.luo@helsinki.fi} \& \href{mailto:shaoxiong.ji@utu.fi}{shaoxiong.ji@utu.fi}
    }
    \small{\textsuperscript{$\dag$} Equal contribution.}
}

\begin{document}
\maketitle
\begin{abstract}

Large language models (LLMs) are advancing at an unprecedented pace globally, with regions increasingly adopting these models for applications in their primary languages. Evaluating these models in diverse linguistic environments, especially in low-resource languages, has become a major challenge for academia and industry. Existing evaluation frameworks suffer from inconsistency across different benchmarks, being disproportionately focused on English and a handful of high-resource languages, thereby overlooking the realistic performance of LLMs in multilingual and lower-resource scenarios. To address this critical challenge of fragmented and inconsistent multilingual evaluation, we introduce GlotEval, a unified and lightweight framework that systematically integrates 27 benchmarks under a standardized ISO 639-3 language identifier system, allowing for seamless incorporation of new benchmarks. Supporting nine key tasks (machine translation, text classification, summarization, open-ended generation, reading comprehension, sequence labeling, intrinsic evaluation, instruction following and reasoning), spanning over dozens to hundreds of languages, GlotEval uniquely enables language-specific, cross-benchmark analysis and non-English-centric evaluations at a scale previously less practical for many researchers. This enables a precise diagnosis of model strengths and weaknesses in diverse linguistic contexts. A multilingual translation case study demonstrates GlotEval's applicability for multilingual and language-specific evaluations.

\vspace{2ex}
\noindent \twemoji{toolbox} GlotEval: \href{https://github.com/MaLA-LM/GlotEval}{github.com/MaLA-LM/GlotEval}\\

\end{abstract}

\section{Introduction}

In recent years, driven by rapid progress in natural language processing and deep learning, large language models (LLMs) such as GPT-4 \citep{openai2023gpt4} and DeepSeek-R1 \citep{deepseekai2025r1} have shown remarkable reasoning and generation capabilities across multiple languages and tasks. Although these models approach or surpass expert-level performance in certain high-resource languages (e.g., English), they often exhibit substantial performance fluctuations in other linguistic environments \citep{zhang2024getting}. This discrepancy partially arises from the imbalance and scarcity of training data of low-resource languages, and partially from the limited multilingual coverage of current evaluation frameworks: many were originally designed for English or a few widely spoken languages, making it difficult to extend them efficiently to more diverse linguistic tasks or to adapt custom prompts and configurations for each language. Meanwhile, as LLMs proliferate worldwide and different regions rely on their respective local languages, large-scale (massively) multilingual evaluation involving numerous low-resource languages has emerged as a critical research direction.

Recent developments in LLM evaluation toolkits such as EleutherAI’s LM Evaluation Harness \citep{eval-harness} and UltraEval \citep{he2024ultraeval} have facilitated automatic evaluation. However, significant gaps persist in language coverage, task diversity, and evaluation flexibility \citep{chang2023survey}, especially in evaluating multilingual LLMs in a massively multilingual scenario.
To address these issues, we present GlotEval, an evaluation framework designed to provide \emph{systematic} support for a \emph{broad range of languages}, with a strong focus on low-resource ones. Building on the core processes of LLM evaluation---data preparation, model inference, post-processing, and metric computation---GlotEval introduces three novel features.

\begin{enumerate}
    \item \textbf{Consistent Cross-benchmark Multilingual Evaluation.}
     We integrate 27 existing multilingual benchmarks into a unified pipeline, by standardizing all ISO 639-3 language codes in the different benchmarks,\footnote{https://iso639-3.sil.org/about} which is an accepted standard with a good coverage of the world's languages. By aligning benchmark language identifiers with ISO 639-3 codes, we enable evaluations for specific languages or language groups (e.g., Bantu, Dravidian, or Uralic languages), allowing the framework to automatically search among integrated benchmarks to find matching test sets. This mapping also makes it easier to incorporate new large-scale benchmarks that target mid- or low-resource languages, ensuring flexibility for future expansions.
    
    \item \textbf{Language-Specific Prompt Templates.} 
    Users can configure prompts for each language individually, thereby enabling more precise assessments of a model’s instruction-following ability across diverse linguistic settings. All templates are maintained in a centralized prompt library that supports multilingual benchmarks, allowing easy customization as needed. In this way, each task within a benchmark can be run potentially using prompts in the task’s original language, rather than defaulting to English prompts. To simplify cross-lingual adaptation, we also implemented Microsoft Translator integration that automatically propagates user-defined prompt templates from one single language to 130+ supported languages.\footnote{https://learn.microsoft.com/en-us/azure/ai-services/translator/language-support}
    
    \item \textbf{Non-English-Centered Machine Translation Evaluation.} 
    GlotEval is designed to break away from the traditional English-centric paradigm. Thanks to translation benchmarks featuring fully or partially multi-aligned datasets, GlotEval enables non-English-centered translation evaluations by allowing any supported language to serve as the pivot: users simply update the pivot language in the configuration to assess  “any-to-pivot” / “pivot-to-any” translation directions. This flexibility ensures that GlotEval breaks from the traditional ``English $\leftrightarrow$ other language'' paradigm and adapts seamlessly to diverse, potentially low-resource, language pairs.

\end{enumerate}

By bringing all these capabilities together in a cohesive framework, 
GlotEval aims to facilitate large-scale, in-depth evaluations of multilingual LLMs 
across both widely spoken and underrepresented languages,
ultimately driving forward more inclusive LLM evaluation.
Thus, GlotEval's primary contribution is not the collection of new tasks, but the synergistic integration and standardization of existing benchmarks, which can be a robust tool for researchers and developers conducting massively multilingual LLM evaluation.

\section{Related Work}

Several evaluation toolkits and benchmarks have been developed to systematically assess LLMs. EleutherAI’s LM Evaluation Harness \citep{eval-harness} is a widely adopted framework covering over 60 tasks, including multilingual datasets such as XNLI (15 languages) and Belebele (122 languages). UltraEval \citep{he2024ultraeval} improves modularity and supports FLORES-200 for multilingual translation. OpenAI Evals provides a highly flexible, community-driven framework,\footnote{https://github.com/openai/evals} and OpenCompass \citep{2023opencompass} offers a comprehensive platform with broad support for datasets and models. MEGA \citep{ahuja-etal-2023-mega} evaluates generative LLMs across diverse languages, with a focus on standard NLP benchmarks.
LightEval \citep{lighteval} developed a flexible LLM evaluation framework that supports different backends.

Despite these advancements, significant gaps remain in language coverage, task diversity, and evaluation flexibility \citep{chang2023survey}. Specifically, most toolkits rely on static task definitions and rarely adopt standardized language identifiers across benchmarks, making it difficult to conduct language-specific evaluations in a cross-benchmark setting. As a result, evaluations for a given language (group) must often be performed in isolation for each benchmark, limiting scalability and linguistic granularity. Furthermore, support for language-specific prompt customization is limited—most toolkits default to using English prompts regardless of the task language, which failed to take both goals of languages in multilingual evaluation, i.e., task performance versus language understanding, into consideration \citep{poelman2024rolesenglishevaluatingmultilingual}.

\section{GlotEval}

\subsection{Benchmarks, Languages and Metrics}

\begin{table*}[t]
\centering
\resizebox{\textwidth}{!}{%
\begin{tabular}{l l l l l l}
\toprule
\textbf{Task} & \textbf{Benchmark} & \textbf{Languages} & \textbf{Domain} & \textbf{Open Source} & \textbf{Metrics} \\ 
\midrule
\multirow{2}{*}{Text Classification} 
  & Taxi-1500 \cite{ma-etal-2024-taxi1500} & 1507 & Bible text & Yes (\href{https://github.com/cisnlp/Taxi1500}{GitHub}) & Acc., F1 \\
  & SIB-200 \cite{adelani-etal-2024-sib} & 205 & News topics & Yes (\href{https://huggingface.co/datasets/cisnlp/SIB-200}{HF}, \href{https://github.com/afrlab-nus/SIB200}{GitHub}) & Acc., F1 \\
\hline
\multirow{2}{*}{Token Classification} 
  & WikiANN \cite{pan-etal-2017-cross} & 282 & Wikipedia NER & Yes (\href{https://huggingface.co/datasets/wikiann}{HF}) & F1 \\
  & UD treebank v2.15 \cite{de-marneffe-etal-2021-universal} & 148 & POS tagging & Yes (\href{https://universaldependencies.org}{UD website}) & F1 \\
\hline
\multirow{11}{*}{Machine Translation} 
  & FLORES-200 \cite{nllb2022} & 200+ & General web & Yes (\href{https://huggingface.co/datasets/facebook/flores}{HF}) & BLEU, ChrF++, COMET \\
  & FLORES+& 212 & Gen. web, low-resource focus & Yes (\href{https://huggingface.co/datasets/facebook/flores}{HF}) & BLEU, ChrF++, COMET \\
  & NTREX-128 \cite{federmann-etal-2022-ntrex} & 128 & News & Yes (\href{https://github.com/facebookresearch/flores/blob/main/ntrex}{GitHub}) & BLEU, ChrF++, COMET \\
  & AmericasNLP \cite{americasnlp-2025-findings} & 14 & Short sentences, court proceedings, books. & Yes (\href{https://github.com/AmericasNLP/americasnlp2021}{GitHub}) & BLEU, ChrF++ \\
  & TICO-19 \cite{anastasopoulos2020tico} & 37 & COVID-19 medical & Yes (\href{https://github.com/tico-19/tico-19.github.io}{GitHub}, \href{https://opus.nlpl.eu/TICO-19.php}{OPUS}) & BLEU, ChrF++ \\
  & IN22 \cite{gala2023indictrans2} & 23 & Indian langs., news+conv. & Yes (\href{https://github.com/AI4Bharat/IndicTrans2}{GitHub}) & BLEU, ChrF++ \\
  & NTEU \cite{bie-etal-2020-neural} & 24 & EU formal (gov) & Partial (Upon request) & BLEU, ChrF++ \\
  & MAFAND \cite{adelani-etal-2022-mafand} & 22 & News & Yes (\href{https://github.com/masakhane-io/masakhane-mt}{GitHub}) & BLEU, ChrF++ \\
  & Tatoeba Challenge v2023 \cite{tiedemann-2020-tatoeba} & 500+ & Mixed short sents. & Yes (\href{https://github.com/Helsinki-NLP/Tatoeba-Challenge}{GitHub}) & BLEU, ChrF \\
& OpenSubtitles v2024 \cite{lison-tiedemann-2016-opensubtitles2016} & 93 & Subtitles & Yes (\href{https://github.com/Helsinki-NLP/OpenSubtitles-devtest}{GitHub})& BLEU, ChrF \\
  & MMHB \cite{tan2024massivemultilingualholisticbias} & 9& Multilingual bias detection & Yes (\href{https://github.com/facebookresearch/ResponsibleNLP/tree/main/mmhb}{GitHub}) & ChrF with gender \\
\hline
\multirow{2}{*}{Open-Ended Generation} 
  & Aya \cite{singh-etal-2024-aya} & 119 & Instruction-following & Yes (\href{https://huggingface.co/datasets/CohereForAI/aya_dataset}{HF}) & self-BLEU \\
  & PolyWrite \cite{ji-etal-2024-emma500} & 240 & Creative writing & Yes (\href{https://huggingface.co/datasets/MaLA-LM/PolyWrite}{HF}) & self-BLEU \\
\hline
\multirow{2}{*}{Intrinsic Evaluation} 
  & PBC \cite{mayer-cysouw-2014-creating} & 372+ & Bible text  & Partial (Upon request) & NLL \\
  & MaLA \citep{ji-etal-2024-emma500} & 546 & General web & Yes (\href{https://huggingface.co/datasets/MaLA-LM/mala-monolingual-split/viewer/default/validation}{HF}) & NLL \\
\hline
\multirow{2}{*}{Comprehension} 
  & MMMLU \cite{hendryckstest2021}& 14+ & General knowledge QA & Yes (\href{https://huggingface.co/datasets/openai/MMMLU}{HF})& Acc. \\
  & Global-MMLU \cite{singh-etal-2024-globalmmlu} & 42 & Culture-aware QA & Yes (\href{https://huggingface.co/datasets/CohereForAI/Global-MMLU}{HF}) & Acc. \\
\hline
\multirow{3}{*}{Summarization} 
  & XLSum \cite{hasan-etal-2021-xl} & 44 & News & Yes (\href{https://huggingface.co/datasets/csebuetnlp/xlsum}{HF}, \href{https://github.com/csebuetnlp/xl-sum}{GitHub}) & ROUGE \\
            & MassiveSumm Long \citep{varab-schluter-2021-massivesumm} & 55 & News & Yes (\href{https://huggingface.co/datasets/MaLA-LM/MassiveSumm_long}{HF}) & ROUGE\\
            & MassiveSumm Short \citep{varab-schluter-2021-massivesumm}& 88 & News & Yes (\href{https://huggingface.co/datasets/MaLA-LM/MassiveSumm_short}{HF}) &  ROUGE \\
\hline
Instruction Following & BenchMAX Rule-based \citep{huang2025benchmaxcomprehensivemultilingualevaluation} & 17 & Verifiable instructions & Yes (\href{https://huggingface.co/datasets/LLaMAX/BenchMAX_Rule-based}{HF}) & Instruction-level Acc. etc. \\
\hline
\multirow{2}{*}{Reasoning} & BenchMAX Math \citep{huang2025benchmaxcomprehensivemultilingualevaluation} & 17 & Grade School Math & Yes (\href{https://huggingface.co/datasets/LLaMAX/BenchMAX_Math}{HF}) & Accuracy \\
 & BenchMAX Science \citep{huang2025benchmaxcomprehensivemultilingualevaluation} & 17 & Graduate-level Scientific QA & Yes (\href{https://huggingface.co/datasets/LLaMAX/BenchMAX_Science}{HF}) & Accuracy \\
\bottomrule
\end{tabular}%
}
\caption{Overview of multilingual LLM evaluation benchmarks, with typical metrics used in each.}
\label{tab:benchmark-overview}
\end{table*}

As shown in Table~\ref{tab:benchmark-overview}, GlotEval integrates publicly available multilingual benchmark datasets, covering machine translation, text classification, summarization, open-ended generation, reading comprehension, sequence labeling, intrinsic evaluation, instruction following and reasoning, spanning a wide range of languages from high-resource to low-resource. In total, GlotEval comprises 9 tasks and 27 benchmarks, evaluates in over 1500 languages, and utilizes diverse metrics. Refer to Appendix~\ref{app:benchmarks} for more details of supported benchmark datasets. 

\subsection{Workflow}

\begin{figure*}[tbph] 
    \centering
    \includegraphics[width=0.95\textwidth]{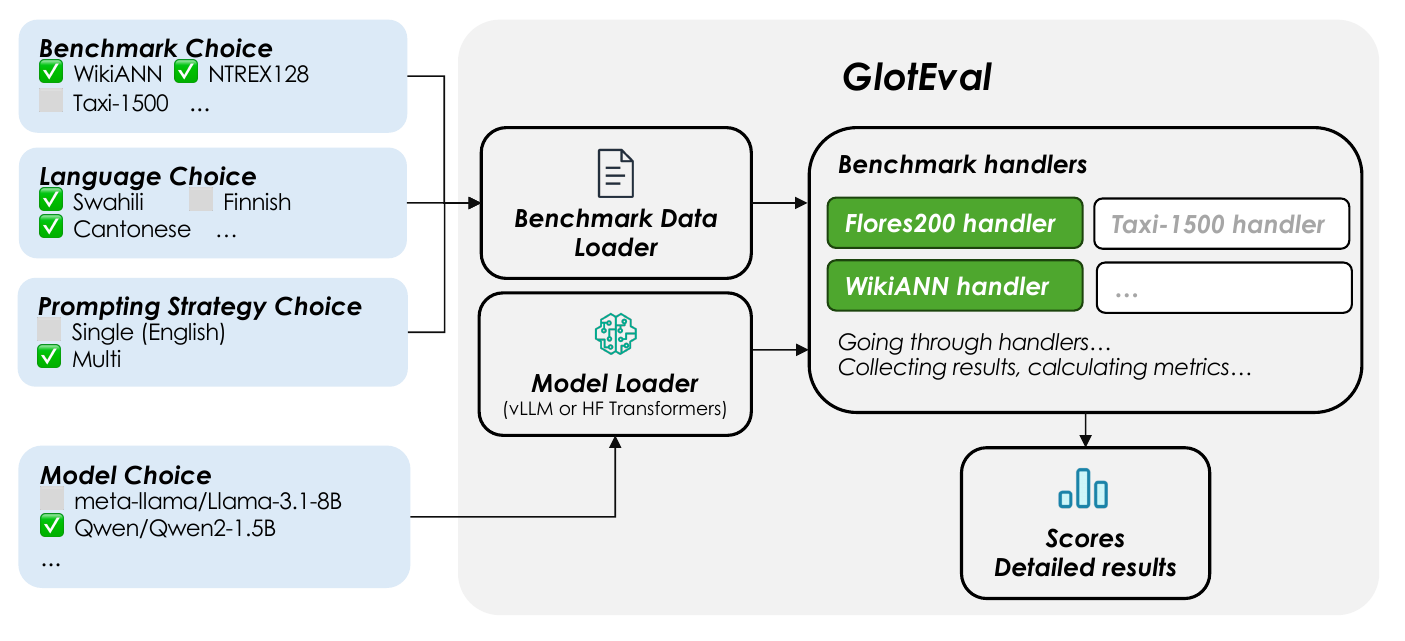} %
    \caption{Workflow of GlotEval}
    \label{fig:workflow}
\end{figure*}

As shown in Figure~\ref{fig:workflow}, the workflow of GlotEval proceeds from specifying which benchmarks and languages to use, to producing final metrics and visualization.

First, users specify their choices 
and through command-line arguments.  Users can specify the language(s) and the benchmark task(s) to evaluate. Besides, as for prompting strategy choice, GlotEval supports two prompting strategies: Setting prompting strategy as \emph{single} along with a chosen prompt language (e.g., \texttt{eng\_Latn}) applies the same prompt in one single language for every dataset in one benchmark. This is useful for controlling variables or using a single reference prompt style; Setting prompting strategy as \emph{multi} makes GlotEval search for a language-specific template in the prompt library, which corresponds to the tested language, falling back to English if not found. Especially in machine translation tasks, the source language typically determines the prompt's language by default. Further, users can freely modify or expand the prompt library with a built-in multilingual prompt builder.

Upon selecting the desired benchmarks, languages, and prompt strategy, the user triggers GlotEval’s data loader to automatically locate each dataset and load the relevant language subsets. It then initializes the appropriate model backend depending on the task type. Specifically, for non-generative tasks, we employ the HuggingFace Transformers backend \citep{wolf-etal-2020-transformers} to ensure more efficient use of computational resources. For generation tasks, such as machine translation, summarization, and open-ended text generation, we prioritize the vLLM backend \citep{kwon2023efficient} to ensure high throughput, while retaining the HF Transformers generation interface for compatibility purposes.

After model inference is completed, GlotEval automatically computes evaluation metrics according to the task-specific settings listed in Table~\ref{tab:benchmark-overview}. Optionally, as mentioned before, by appending \texttt{--store\_details}, users can export each sample’s prompt, model output, reference, and corresponding scores to a JSONL file, which allows researchers to work outside the framework and conduct custom error analysis and result visualization. This ensures that our framework is not just an evaluation executor, but also a starting point for more fine-grained analysis.

\subsection{A Deeper Look at Benchmark Data Loader}

\begin{figure*}[tbph] 
    \centering
    \includegraphics[width=0.95\textwidth]{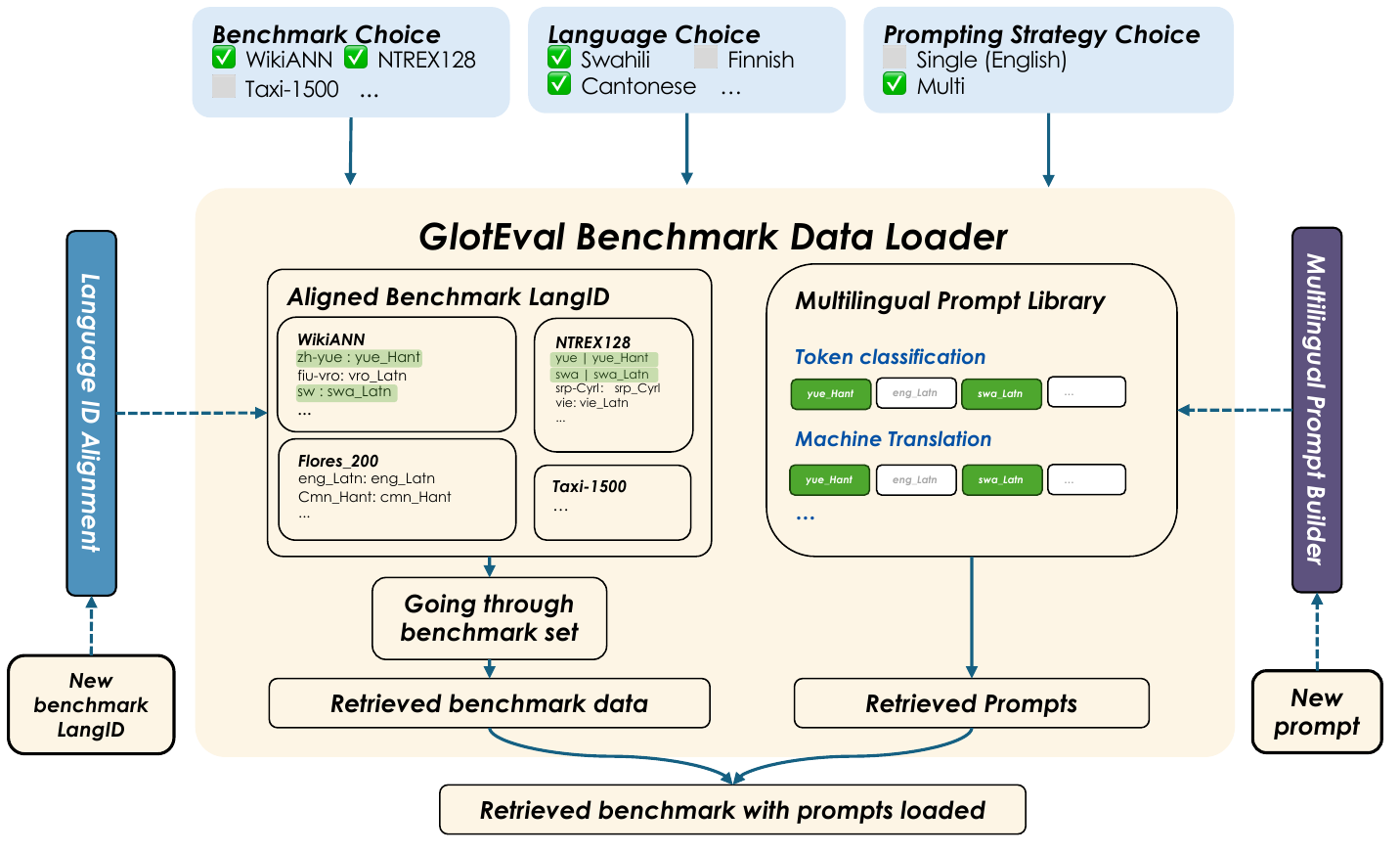} %
    \caption{GlotEval benchmark data loader}
    \label{fig:data_loader}
\end{figure*}

Figure~\ref{fig:data_loader} illustrates the overall workflow within GlotEval's data loading and prompt preparation pipeline. At its core, GlotEval aligns language identifiers from various benchmarks to a unified \texttt{ISO~639-3\_Script} format. Once the alignment is complete, the standardized language codes serve as the central connection for downstream operations. When the user queries GlotEval with a target language (e.g., \texttt{zho} for Chinese or \texttt{spa} for Spanish), the system consults the language-to-code dictionary and retrieves all benchmark-specific subsets whose original language codes map to the same standardized form. These subsets are then included in the evaluation process. Moreover, if a language-specific prompting strategy is selected, GlotEval uses the same aligned codes to retrieve the appropriate prompt templates from the multilingual prompt library. For example, as shown in Figure~\ref{fig:data_loader}, querying \texttt{zho} and \texttt{spa} will automatically select the corresponding benchmark subsets and load their respective prompts (\texttt{zho\_Hans}, \texttt{spa\_Latn}) for evaluation. This workflow builds on both the language code alignment mechanism and the multilingual prompt builder described in the following sections.

\subsubsection*{Language Code Alignment to ISO 639-3}

Different benchmarks often use inconsistent codes for the same language (e.g., \texttt{zh}, \texttt{zho}, \texttt{cmn}, \texttt{Chinese}, \texttt{Mandarin-CN} etc. for Mandarin Chinese). Before reading benchmark datasets via dedicated data loaders,  GlotEval unifies these language identifiers used across different benchmarks, to enable cross-benchmark language-specific evaluation and prompting. Figure~\ref{fig:component_a} visualizes this process.

Specifically, we process each benchmark-provided language code—which may appear in the form of ISO 639-1, 639-2/B (bibliographic), 639-2/T (terminological), ISO 639-3 codes, or even language names—by utilizing the \texttt{iso639-lang} Python package.\footnote{https://pypi.org/project/iso639-lang/}  This allows us to retrieve all available mappings from the ISO 639-3 standard, including ISO 639-3 identifiers, ISO 639-2/B, 639-2/T, and ISO 639-1 codes. Using both exact and fuzzy matching strategies, we attempt to automatically identify the corresponding ISO 639-3 code for each language. A report is generated that documents, for each benchmark language, whether the match was exact or fuzzy, and whether it corresponds to an individual language or a macrolanguage in the ISO 639-3 standard.

We further identify the script used by each dataset, using \texttt{GlotScript} \citep{kargaran-etal-2024-glotscript} to detect the dominant script.\footnote{https://pypi.org/project/GlotScript/} Here we assume each dataset is primarily in one script. We randomly select up to 100 lines and attempt script recognition into ISO 15924 script code. This ensures each dataset obtains a \texttt{\textless language\textgreater\_\textless script\textgreater} label, such as \texttt{eng\_Latn}. The final ISO 639-3 code, along with the script code, is stored as the value in a language-to-code dictionary within the benchmark’s configuration file. Hence, each language + script combination is standardized in GlotEval for consistent usage across benchmarks.

\subsubsection*{Multilingual Prompt Builder}

\begin{figure}[tbph]
    \centering
    \begin{subfigure}[b]{0.48\textwidth}
        \centering
        \includegraphics[width=\textwidth]{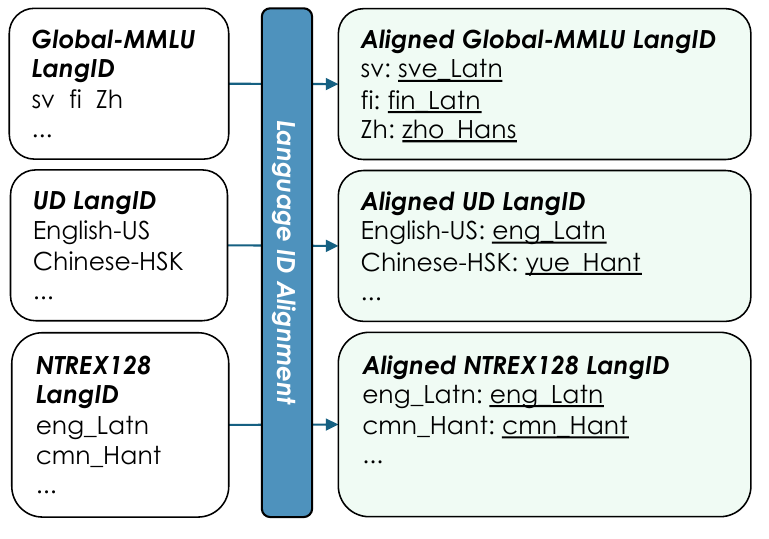}
        \caption{}
        \label{fig:component_a}
    \end{subfigure}
    
    \vspace{0.5em} 

    \begin{subfigure}[b]{0.48\textwidth}
        \centering
        \includegraphics[width=\textwidth]{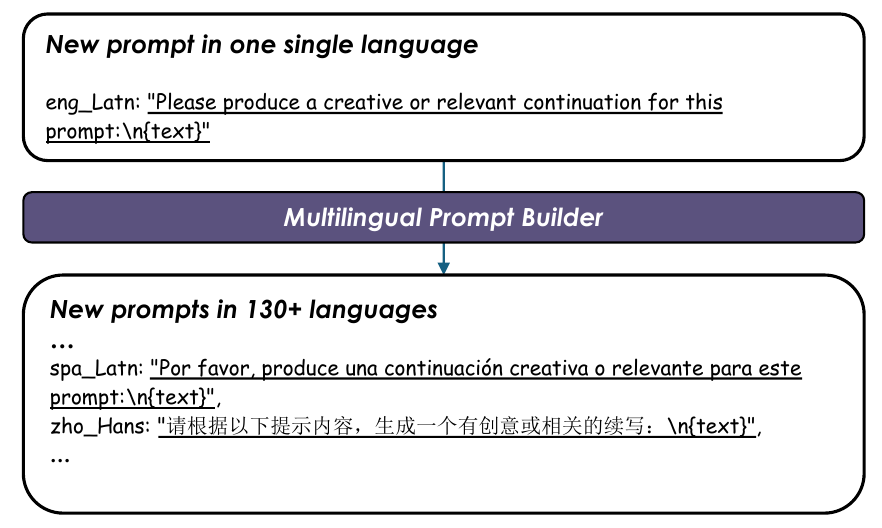}
        \caption{}
        \label{fig:component_b}
    \end{subfigure}
    
    \caption{Benchmark data loader components: (a) Language ID alignment process and (b) multilingual prompt generation.}
    \label{fig:benchmark_components}
\end{figure}

We constructed a dedicated command-line prompt builder to automatically prepare or adapt prompt templates for multilingual tasks. Figure~\ref{fig:component_b} visualizes this process. 
In particular, the builder leverages Microsoft Translator to convert an instruction and/or few-shot prompt template from a given source language into 130+ target languages, while ensuring that placeholders (e.g., \texttt{\{src\_text\}}) remain intact during translation.
These newly created multilingual prompts, are stored in the updated prompt library. As a result, each dataset’s prompts are aligned with the same \texttt{\textless language\textgreater\_\textless script\textgreater} language taxonomy, enabling consistent, language-specific evaluation. 

Note that the automatic translation of prompts is intended as a convenience feature to support rapid, large-scale multilingual evaluation. While translation quality may vary—particularly for low-resource languages—this approach offers a practical starting point for exploratory analysis with language-specific prompts at scale. The framework remains fully customizable: users are able to provide their own human-written or verified prompts in the prompt library for languages of interest.

\section{Evaluation}

\subsection{Efficiency Analysis}

\label{sec:efficiency}

We benchmark GlotEval’s inference speed on six tasks:
FLORES-200, Aya, and XLSum for generative tasks, and
SIB-200, Global-MMLU, and WikiANN for non-generative tasks.
All evaluations are conducted on 19 languages spanning diverse writing systems (e.g., Latin, Arabic, Cyrillic, Devanagari, Chinese, etc.). For each language, we sample 10 examples per task for evaluation. We choose Qwen2-1.5B model \cite{qwen2} for evaluation.
For generative tasks, we measure generation throughput (prefilling and decoding) with vLLM backend.
For non-generative tasks, we measure classification throughput (prefilling only) using HF Transformers.

We consider two GPU environments:
\begin{itemize}[noitemsep,nolistsep]
    \item \textbf{AMD MI250X 64GB} (BF16, single GPU, batch size set as 1)
    \item \textbf{NVIDIA A100 40GB} (BF16, single GPU, batch size set as 1)
\end{itemize}

For detailed throughput performance, Appendix~\ref{app:efficiency} shows statistics on both GPU environments. They demonstrate that in general, NVIDIA A100 consistently achieves higher throughput than AMD MI250X across both generative and non-generative tasks. Besides, this gap may also reflect the different backends between vLLM and HF Transformers. We further observe that scripts such as Devanagari or Amharic (\texttt{amh\_Ethi}) often have lower throughput, potentially due to more complex tokenization. Lastly, summarization tasks like XLSum typically involve longer inputs and outputs than sentence-level translation tasks (e.g., FLORES-200), which increases the prefilling overhead and thus reduces the overall tokens/s.

\subsection{Case Study on Multilingual Translation}

\begin{table*}[]
\centering
\resizebox{0.95\textwidth}{!}{%
\begin{tabular}{lllll}
\cline{1-3}
Prompt Strategy                                                   & Tested Translation Language Pair                                                                      & Prompt Template                                                                                                                                                                                                         &  &  \\ \cline{1-3}
\begin{tabular}[c]{@{}l@{}} \texttt{multi}\\ language-specific\end{tabular} & \begin{tabular}[c]{@{}l@{}} \texttt{fra → fin}\\ French → Finnish\end{tabular}                      & \begin{tabular}[c]{@{}l@{}}Traduisez la phrase suivante de \textbf{Langue française} en \textbf{Langue finnoise}\\ {[}Langue française{]} : \{source\_text\_in\_finnish\}\\ {[}Langue finnoise{]} :\end{tabular}                          &  &  \\ \cline{1-3}
\begin{tabular}[c]{@{}l@{}}\texttt{fin\_Latn}\\ Finnish\end{tabular}       & \begin{tabular}[c]{@{}l@{}}\texttt{vie → zho-CN}\\ Vietnamese → Chinese (Simplified)\end{tabular} & \begin{tabular}[c]{@{}l@{}}Käännä seuraava lause \textbf{Vietnamin kieli} muotoon \textbf{Kiinan kieli (yksinkertaistettu)}\\ {[}Vietnamin kieli{]}: \{source\_text\_in\_vietnamese\}\\ {[}Kiinan kieli (yksinkertaistettu){]}:\end{tabular} &  &  \\ \cline{1-3}
\end{tabular}%
}
\caption{A demonstration of prompt templates of translation tasks in different prompt strategies.}
\label{tab:prompt-overview}
\end{table*}

To further illustrate GlotEval's capabilities, we conducted a detailed case study comparing EMMA-500 \citep{ji-etal-2024-emma500}, a large-scale multilingual language model designed to enhance multilingual performance, with the base Llama-2-7B model \citep{llama2} across various multilingual translation scenarios. This study aimed to investigate performance differences under different prompting strategies and diverse language-centric translation tasks. We designed a factorial experiment with the following variables:

\begin{itemize}[nolistsep,noitemsep]
    \item \textbf{Models}: EMMA-500 vs. Llama-2-7B
    \item \textbf{Prompting strategies}: multilingual prompting (source language-specific), Chinese prompting (zho\_Hans), Finnish prompting (fin\_Latn), and English prompting (eng\_Latn)
    \item \textbf{Translation directions}: six configurations with different central languages (X$\rightarrow$eng-US, eng-US$\rightarrow$X, zho-CN$\rightarrow$X, X$\rightarrow$zho-CN, fin$\rightarrow$X, X$\rightarrow$fin)
\end{itemize}

A demonstration of prompt templates is shown in Table~\ref{tab:prompt-overview}. For evaluation, we utilized NTREX-128, a multi-aligned benchmark containing parallel texts across 128 languages, which is supported in GlotEval. In the multilingual prompting condition, we used built-in prompt builder in GlotEval, with the support of Microsoft Translator service, to automatically translate prompts into 134 languages supported by their platform. In our case study, 106 of these languages overlap with NTREX-128 languages, allowing us to test performance across this diverse language set. 

\begin{figure}[tbph] 
    \centering
    \includegraphics[width=0.47\textwidth]{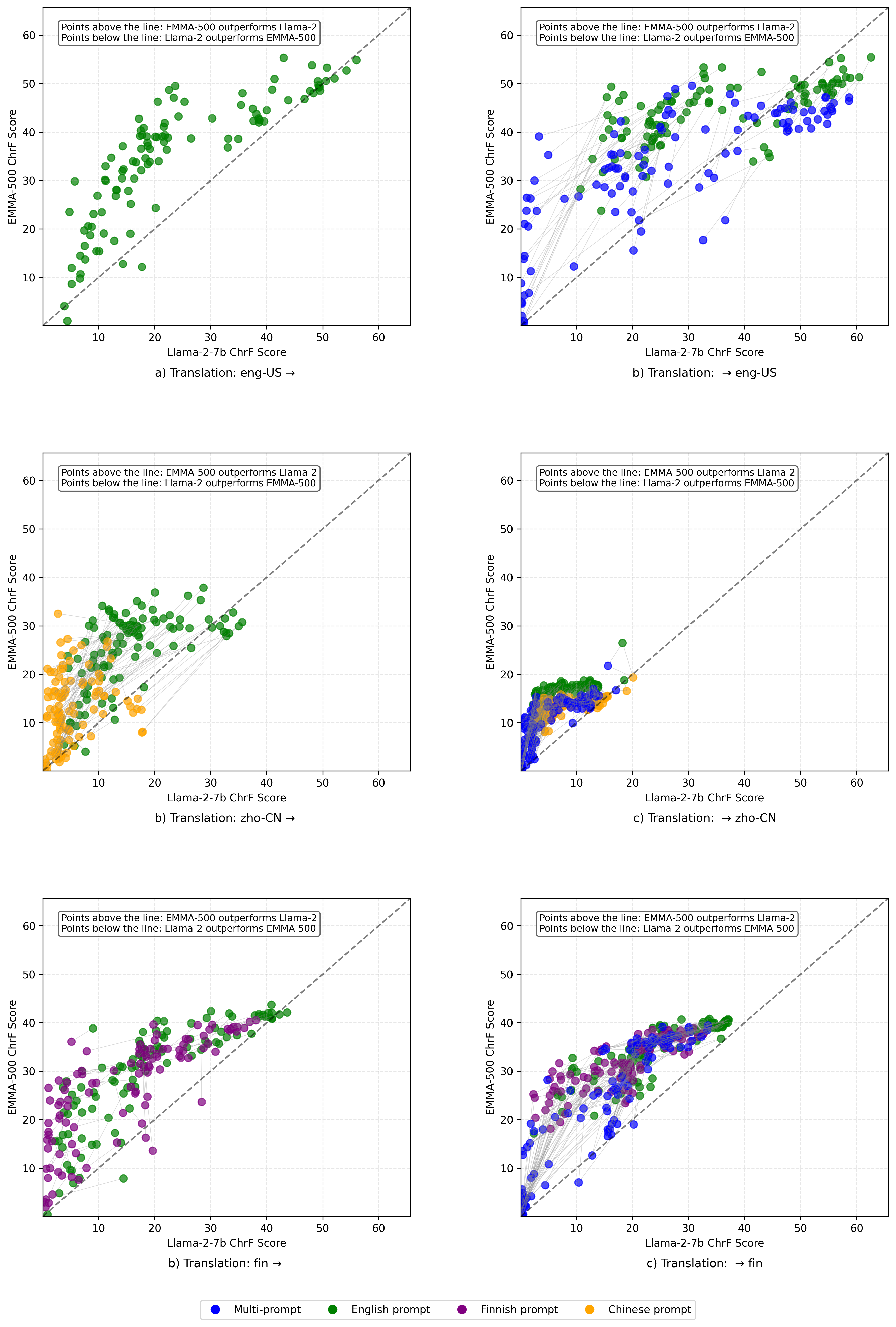} %
    \caption{ChrF scores for different translation directions comparing EMMA-500 and Llama-2-7B across four prompting strategies.}
    \label{fig:translation-results}
\end{figure}

The results of our case study (Figure~\ref{fig:translation-results}) clearly demonstrate EMMA-500's performance compared to Llama-2-7B in multilingual instruction following capabilities and non-English-centric translation tasks. Specifically, EMMA-500 shows consistently higher ChrF scores across most language pairs for all six translation directions. This performance advantage is particularly pronounced when using non-English prompting strategies, highlighting EMMA-500's enhanced ability to process and respond to instructions in diverse languages.

The experimental design was implemented using GlotEval, which facilitated the systematic manipulation of variables through simple configuration settings. By simply modifying the prompting strategy parameter and central language settings in the multi-aligned MT benchmark configuration, we are able to comprehensively assess the language models' multilingual capabilities, including both instruction following and non-English-centric multilingual translation.

\section{Conclusion and Future Work}

In this work, we introduced GlotEval, a lightweight yet comprehensive framework for massively multilingual evaluation of LLMs. By supporting consistent multilingual benchmarking, incorporating language-specific prompt templates, and supporting flexible non-English-centric translation setups, GlotEval enables consistent assessments of LLMs in diverse linguistic contexts—including low-resource settings often neglected by traditional benchmarks. Our case study on multilingual machine translation with two LLMs illustrates the utility of GlotEval in revealing the strengths and weaknesses of multilingual LLMs and in identifying directions for future optimization. Overall, GlotEval aims to encourage more inclusive, transparent, and holistic evaluations of language models across a wide array of languages and tasks, thereby advancing robust multilingual NLP research.

As for future work, we plan to integrate more diverse and comprehensive multilingual benchmarks to better evaluate LLM performance. Plus, we will explore the integration of benchmarks that  the synergistic combination of automatic and human evaluation; for example, this could achieved through our pilot development of a lightweight web interface that supports crowd-sourced and expert-driven evaluation to supplement the automatic evaluation.\footnote{Source code and documentation are available at \url{https://github.com/MaLA-LM/GlotEval-HumanEval} and \url{https://gloteval-humaneval.readthedocs.io}}

\section*{Ethical Considerations and Broader Impact}
\paragraph{Ethical Considerations} We strive to uphold the principles outlined in the \href{https://www.aclweb.org/portal/content/acl-code-ethics}{ACL Code of Ethics}.
While GlotEval advances multilingual evaluation, several limitations remain. Many benchmarks still lack sufficient or high-quality data for truly low-resource languages, potentially skewing performance assessments. Additionally, as noted by \citet{Aditya2025NLP4dialects}, existing datasets often inherit cultural and linguistic biases, favoring dominant dialects or standardized language forms over regional or marginalized variants. Computational costs further constrain accessibility: large-scale evaluations are resource-intensive, posing barriers for smaller research teams. More critically, reference-free metrics introduce inherent biases, as they effectively pit one generative model against another \citep{deutsch-etal-2022-limitations}. Such metrics struggle to capture fluency, accuracy, or cultural appropriateness, particularly in low-resource contexts where human judgments are essential.

\paragraph{Broader Impact} GlotEval promotes equitable progress in NLP by enabling systematic evaluation of large language models (LLMs) across diverse languages. We aim to support researchers and developers in creating language technologies that serve diverse communities more effectively via a more inclusive and holistic evaluation suite.

\section*{Acknowledgments}

This project is funded by the AI-DOC program hosted by Finnish Center of Artificial Intelligence (decision number VN/3137/2024-OKM-6).

The work has received funding from the European Union's Horizon Europe research and innovation programme under grant agreement No 101070350 and from UK Research and Innovation (UKRI) under the UK government's Horizon Europe funding guarantee [grant number 10052546], and the Digital Europe Programme under grant agreement No 101195233.

The authors wish to acknowledge CSC - IT Center for Science, Finland, the Leonardo and LUMI supercomputers, owned by the EuroHPC Joint Undertaking, for providing computational resources.

Sawal Devkota, Bhavani Sai Praneeth Varma Mantina, Ananda Sreenidhi, Mengjie Wang, and Samea Yusofi contributed to this project as part of the ``Data Analysis Software Project for Natural Language'' course at TU Darmstadt, under the guidance of Shaoxiong Ji. This teaching activity was funded by LOEWE Center DYNAMIC as part of the Hessian program for the promotion of cutting-edge research LOEWE under the grant number of LOEWE1/16/519/03/09.001(0009)/98.

\bibliography{gloteval}

\begin{thebibliography}{52}
\providecommand{\natexlab}[1]{#1}

\bibitem[{Abdulmumin et~al.(2024)Abdulmumin, Mkhwanazi, Mbooi, Muhammad, Ahmad, Putini, Mathebula, Shingange, Gwadabe, and Marivate}]{wmt24-4african}
Idris Abdulmumin, Sthembiso Mkhwanazi, Mahlatse~S. Mbooi, Shamsuddeen~Hassan Muhammad, Ibrahim~Said Ahmad, Neo~N. Putini, Miehleketo Mathebula, Matimba Shingange, Tajuddeen Gwadabe, and Vukosi Marivate. 2024.
\newblock Correcting {FLORES} evaluation dataset for four {African} languages.
\newblock In \emph{Proceedings of the Ninth Conference on Machine Translation}, Miami, USA. Association for Computational Linguistics.

\bibitem[{Adelani et~al.(2022)Adelani, Alabi, Fan, Kreutzer, Shen, Reid, Ruiter, Klakow, Nabende, Chang et~al.}]{adelani-etal-2022-mafand}
David Adelani, Jesujoba Alabi, Angela Fan, Julia Kreutzer, Xiaoyu Shen, Machel Reid, Dana Ruiter, Dietrich Klakow, Peter Nabende, Ernie Chang, et~al. 2022.
\newblock A few thousand translations go a long way! leveraging pre-trained models for african news translation.
\newblock pages 3053--3070.

\bibitem[{Adelani et~al.(2024)Adelani, Liu, Shen, Vassilyev, Alabi, Mao, Gao, and Lee}]{adelani-etal-2024-sib}
David~Ifeoluwa Adelani, Hannah Liu, Xiaoyu Shen, Nikita Vassilyev, Jesujoba~O. Alabi, Yanke Mao, Haonan Gao, and En-Shiun~Annie Lee. 2024.
\newblock {SIB}-200: A simple, inclusive, and big evaluation dataset for topic classification in 200+ languages and dialects.
\newblock In \emph{Proceedings of the 18th Conference of the European Chapter of the Association for Computational Linguistics (Volume 1: Long Papers)}, pages 226--245, St. Julian’s, Malta. Association for Computational Linguistics.

\bibitem[{Ahuja et~al.(2023)Ahuja, Diddee, Hada, Ochieng, Ramesh, Jain, Nambi, Ganu, Segal, Ahmed, Bali, and Sitaram}]{ahuja-etal-2023-mega}
Kabir Ahuja, Harshita Diddee, Rishav Hada, Millicent Ochieng, Krithika Ramesh, Prachi Jain, Akshay Nambi, Tanuja Ganu, Sameer Segal, Mohamed Ahmed, Kalika Bali, and Sunayana Sitaram. 2023.
\newblock \href {https://doi.org/10.18653/v1/2023.emnlp-main.258} {{MEGA}: Multilingual evaluation of generative {AI}}.
\newblock In \emph{Proceedings of the 2023 Conference on Empirical Methods in Natural Language Processing}, pages 4232--4267, Singapore. Association for Computational Linguistics.

\bibitem[{AI4Bharat et~al.(2023)AI4Bharat, Gala, Chitale, AK, Doddapaneni, Gumma, Kumar, Nawale, Sujatha, Puduppully, Raghavan, Kumar, Khapra, Dabre, and Kunchukuttan}]{indictrans2-23}
AI4Bharat, Jay Gala, Pranjal~A. Chitale, Raghavan AK, Sumanth Doddapaneni, Varun Gumma, Aswanth Kumar, Janki Nawale, Anupama Sujatha, Ratish Puduppully, Vivek Raghavan, Pratyush Kumar, Mitesh~M. Khapra, Raj Dabre, and Anoop Kunchukuttan. 2023.
\newblock \href {https://arxiv.org/abs/arXiv:2305.16307} {Indictrans2: Towards high-quality and accessible machine translation models for all 22 scheduled indian languages}.

\bibitem[{Ali et~al.(2024)Ali, Cardoso, and Sousa-Silva}]{wmt24-emakhuwa}
Felermino Dario~Mario Ali, Henrique~Lopes Cardoso, and Rui Sousa-Silva. 2024.
\newblock Expanding {FLORES+} benchmark for more low-resource settings: {Portuguese-Emakhuwa} machine translation evaluation.
\newblock In \emph{Proceedings of the Ninth Conference on Machine Translation}, Miami, USA. Association for Computational Linguistics.

\bibitem[{Anastasopoulos et~al.(2020)Anastasopoulos, Cattelan, Dou, Federico, Federmann, Genzel, Guzm{\'a}n, Hu, Hughes, Koehn et~al.}]{anastasopoulos2020tico}
Antonios Anastasopoulos, Alessandro Cattelan, Zi-Yi Dou, Marcello Federico, Christian Federmann, Dmitriy Genzel, Franscisco Guzm{\'a}n, Junjie Hu, Macduff Hughes, Philipp Koehn, et~al. 2020.
\newblock Tico-19: the translation initiative for covid-19.
\newblock In \emph{Proceedings of the 1st Workshop on NLP for COVID-19 (Part 2) at EMNLP 2020}.

\bibitem[{Bi{\'e} et~al.(2020)Bi{\'e}, Cerd{\`a}-i Cuc{\'o}, Degroote, Estela, Garc{\'i}a-Mart{\'i}nez, Herranz, Kohan, Melero, O{'}Dowd, O{'}Gorman, Pinnis, Rozis, Superbo, and Vasi{\c{l}}evskis}]{bie-etal-2020-neural}
Laurent Bi{\'e}, Aleix Cerd{\`a}-i Cuc{\'o}, Hans Degroote, Amando Estela, Mercedes Garc{\'i}a-Mart{\'i}nez, Manuel Herranz, Alejandro Kohan, Maite Melero, Tony O{'}Dowd, Sin{\'e}ad O{'}Gorman, M{\={a}}rcis Pinnis, Roberts Rozis, Riccardo Superbo, and Art{\={u}}rs Vasi{\c{l}}evskis. 2020.
\newblock \href {https://aclanthology.org/2020.eamt-1.60/} {Neural translation for the {E}uropean {U}nion ({NTEU}) project}.
\newblock In \emph{Proceedings of the 22nd Annual Conference of the European Association for Machine Translation}, pages 477--478, Lisboa, Portugal. European Association for Machine Translation.

\bibitem[{Chang et~al.(2024)Chang, Wang, Wang, Wu, Zhu, Chen, Yang, Yi, Wang, Wang et~al.}]{chang2023survey}
Yupeng Chang, Xu~Wang, Jindong Wang, Yuan Wu, Kaijie Zhu, Hao Chen, Linyi Yang, Xiaoyuan Yi, Cunxiang Wang, Yidong Wang, et~al. 2024.
\newblock \href {https://doi.org/10.1145/3641289} {A survey on evaluation of large language models}.
\newblock \emph{ACM Trans. Intell. Syst. Technol.}, 15(3).

\bibitem[{Contributors(2023)}]{2023opencompass}
OpenCompass Contributors. 2023.
\newblock Opencompass: A universal evaluation platform for foundation models.
\newblock \url{https://github.com/open-compass/opencompass}.

\bibitem[{de~Gibert et~al.(2025)de~Gibert, Pugh, Marashian, Vazquez, Ebrahimi, Denisov, Rice, Gow-Smith, Prieto, Robles, Manrique, Moreno~Veliz, Lino~Campos, Coto-Solano, Alvarez, Agüero-Torales, Ortega, Chiruzzo, Oncevay, Rijhwani, von~der Wense, and Mager}]{americasnlp-2025-findings}
Ona de~Gibert, Robert Pugh, Ali Marashian, Raul Vazquez, Abteen Ebrahimi, Pavel Denisov, Enora Rice, Edward Gow-Smith, Juan~C. Prieto, Melissa Robles, Rubén Manrique, Oscar Moreno~Veliz, Ángel Lino~Campos, Rolando Coto-Solano, Aldo Alvarez, Marvin Agüero-Torales, John~E. Ortega, Luis Chiruzzo, Arturo Oncevay, Shruti Rijhwani, Katharina von~der Wense, and Manuel Mager. 2025.
\newblock {F}indings of the {A}mericas{NLP} 2025 {S}hared {T}asks on {M}achine {T}ranslation, {C}reation of {E}ducational {M}aterial, and {T}ranslation {M}etrics for {I}ndigenous {L}anguages of the {A}mericas.
\newblock In \emph{Proceedings of the 5th Workshop on Natural Language Processing for Indigenous Languages of the Americas (AmericasNLP 2025)}, Albuquerque, New Mexico. Association for Computational Linguistics.

\bibitem[{de~Marneffe et~al.(2021)de~Marneffe, Manning, Nivre, and Zeman}]{de-marneffe-etal-2021-universal}
Marie-Catherine de~Marneffe, Christopher~D. Manning, Joakim Nivre, and Daniel Zeman. 2021.
\newblock \href {https://doi.org/10.1162/coli_a_00402} {Universal dependencies}.
\newblock \emph{Computational Linguistics}, 47(2):255--308.

\bibitem[{DeepSeek-AI et~al.(2025)DeepSeek-AI, Guo, Yang, Zhang et~al.}]{deepseekai2025r1}
DeepSeek-AI, Daya Guo, Dejian Yang, Haowei Zhang, et~al. 2025.
\newblock Deepseek-r1: Incentivizing reasoning capability in llms via reinforcement learning.
\newblock \emph{arXiv preprint arXiv:2501.12948}.

\bibitem[{Deutsch et~al.(2022)Deutsch, Dror, and Roth}]{deutsch-etal-2022-limitations}
Daniel Deutsch, Rotem Dror, and Dan Roth. 2022.
\newblock \href {https://doi.org/10.18653/v1/2022.emnlp-main.753} {On the limitations of reference-free evaluations of generated text}.
\newblock In \emph{Proceedings of the 2022 Conference on Empirical Methods in Natural Language Processing}, pages 10960--10977, Abu Dhabi, United Arab Emirates. Association for Computational Linguistics.

\bibitem[{Doumbouya et~al.(2023)Doumbouya, Diané, Cissé, Diané, Sow, Doumbouya, Bangoura, Bayo, Condé, Diané, Piech, and Manning}]{mt4nko-23}
Moussa Doumbouya, Baba~Mamadi Diané, Solo~Farabado Cissé, Djibrila Diané, Abdoulaye Sow, Séré~Moussa Doumbouya, Daouda Bangoura, Fodé~Moriba Bayo, Ibrahima Sory~2. Condé, Kalo~Mory Diané, Chris Piech, and Christopher Manning. 2023.
\newblock \href {https://aclanthology.org/2023.wmt-1.34} {Machine translation for nko: Tools, corpora, and baseline results}.
\newblock In \emph{Proceedings of the Eighth Conference on Machine Translation}, pages 312--343, Singapore. Association for Computational Linguistics.

\bibitem[{Federmann et~al.(2022)Federmann, Kocmi, and Xin}]{federmann-etal-2022-ntrex}
Christian Federmann, Tom Kocmi, and Ying Xin. 2022.
\newblock {NTREX}-128 -- news test references for {MT} evaluation of 128 languages.
\newblock In \emph{Proceedings of the First Workshop on Scaling Up Multilingual Evaluation}, pages 21--24, Online. Association for Computational Linguistics.

\bibitem[{Fourrier et~al.(2023)Fourrier, Habib, Kydlíček, Wolf, and Tunstall}]{lighteval}
Clémentine Fourrier, Nathan Habib, Hynek Kydlíček, Thomas Wolf, and Lewis Tunstall. 2023.
\newblock \href {https://github.com/huggingface/lighteval} {Lighteval: A lightweight framework for llm evaluation}.

\bibitem[{Gala et~al.(2023)Gala, Chitale, Raghavan, Gumma, Doddapaneni, Aswanth, Nawale, Sujatha, Puduppully, Raghavan et~al.}]{gala2023indictrans2}
Jay~P Gala, Pranjal~A Chitale, AK~Raghavan, Varun Gumma, Sumanth Doddapaneni, Kumar~M Aswanth, Janki~Atul Nawale, Anupama Sujatha, Ratish Puduppully, Vivek Raghavan, et~al. 2023.
\newblock Indictrans2: Towards high-quality and accessible machine translation models for all 22 scheduled indian languages.
\newblock \emph{Transactions on Machine Learning Research}, 2023.

\bibitem[{Gao et~al.(2023)Gao, Tow, Biderman, Black, DiPofi, Foster, Golding, Hsu, McDonell, Muennighoff et~al.}]{eval-harness}
Leo Gao, Jonathan Tow, Stella Biderman, Sid Black, Anthony DiPofi, Charles Foster, Laurence Golding, Jeffrey Hsu, Kyle McDonell, Niklas Muennighoff, et~al. 2023.
\newblock A framework for few-shot language model evaluation.
\newblock Zenodo.

\bibitem[{Gordeev et~al.(2024)Gordeev, Kuldin, and Dale}]{wmt24-erzya}
Isai Gordeev, Sergey Kuldin, and David Dale. 2024.
\newblock Flores+ translation and machine translation evaluation for the {E}rzya language.
\newblock In \emph{Proceedings of the Ninth Conference on Machine Translation}, Miami, USA. Association for Computational Linguistics.

\bibitem[{Goyal et~al.(2022)Goyal, Gao, Chaudhary, Chen, Wenzek, Ju, Krishnan, Ranzato, Guzmán, and Fan}]{flores101-22}
Naman Goyal, Cynthia Gao, Vishrav Chaudhary, Peng-Jen Chen, Guillaume Wenzek, Da~Ju, Sanjana Krishnan, Marc’Aurelio Ranzato, Francisco Guzmán, and Angela Fan. 2022.
\newblock The {F}lores-101 evaluation benchmark for low-resource and multilingual machine translation.
\newblock \emph{Transactions of the Association for Computational Linguistics}, 10.

\bibitem[{Guzmán et~al.(2019)Guzmán, Chen, Ott, Pino, Lample, Koehn, Chaudhary, and Ranzato}]{flores1-19}
Francisco Guzmán, Peng-Jen Chen, Myle Ott, Juan Pino, Guillaume Lample, Philipp Koehn, Vishrav Chaudhary, and Marc’Aurelio Ranzato. 2019.
\newblock \href {https://aclanthology.org/D19-1632} {The {FLORES} evaluation datasets for low-resource machine translation: {N}epali{--}{E}nglish and {S}inhala{--}{E}nglish}.
\newblock In \emph{Proceedings of the 2019 Conference on Empirical Methods in Natural Language Processing and the 9th International Joint Conference on Natural Language Processing (EMNLP-IJCNLP)}, pages 6098--6111, Hong Kong, China. Association for Computational Linguistics.

\bibitem[{Hasan et~al.(2021)Hasan, Bhattacharjee, Islam, Mubasshir, Li, Kang, Rahman, and Shahriyar}]{hasan-etal-2021-xl}
Tahmid Hasan, Abhik Bhattacharjee, Md.~Saiful Islam, Kazi Mubasshir, Yuan-Fang Li, Yong-Bin Kang, M.~Sohel Rahman, and Rifat Shahriyar. 2021.
\newblock {XL}-sum: Large-scale multilingual abstractive summarization for 44 languages.
\newblock In \emph{Findings of the Association for Computational Linguistics: ACL-IJCNLP 2021}, pages 4693--4703, Online. Association for Computational Linguistics.

\bibitem[{He et~al.(2024)He, Luo, Hu, Zhao, Zhou, Wu, Zhang, Han, Liu, and Sun}]{he2024ultraeval}
Chaoqun He, Renjie Luo, Shengding Hu, Yuanqian Zhao, Jie Zhou, Hanghao Wu, Jiajie Zhang, Xu~Han, Zhiyuan Liu, and Maosong Sun. 2024.
\newblock \href {https://arxiv.org/abs/2404.07584} {Ultraeval: A lightweight platform for flexible and comprehensive evaluation for llms}.
\newblock \emph{Preprint}, arXiv:2404.07584.

\bibitem[{Hendrycks et~al.(2021)Hendrycks, Burns, Basart, Zou, Mazeika, Song, and Steinhardt}]{hendryckstest2021}
Dan Hendrycks, Collin Burns, Steven Basart, Andy Zou, Mantas Mazeika, Dawn Song, and Jacob Steinhardt. 2021.
\newblock Measuring massive multitask language understanding.
\newblock \emph{Proceedings of the International Conference on Learning Representations (ICLR)}.

\bibitem[{Huang et~al.(2025)Huang, Zhu, Hu, He, Li, Huang, and Yuan}]{huang2025benchmaxcomprehensivemultilingualevaluation}
Xu~Huang, Wenhao Zhu, Hanxu Hu, Conghui He, Lei Li, Shujian Huang, and Fei Yuan. 2025.
\newblock \href {https://arxiv.org/abs/2502.07346} {Benchmax: A comprehensive multilingual evaluation suite for large language models}.
\newblock \emph{Preprint}, arXiv:2502.07346.

\bibitem[{Ji et~al.(2024)Ji, Li, Paul, Paavola, Lin, Chen, O'Brien, Luo, Sch{\"u}tze, Tiedemann, and Haddow}]{ji-etal-2024-emma500}
Shaoxiong Ji, Zihao Li, Indraneil Paul, Jaakko Paavola, Peiqin Lin, Pinzhen Chen, Dayy{\'a}n O'Brien, Hengyu Luo, Hinrich Sch{\"u}tze, J{\"o}rg Tiedemann, and Barry Haddow. 2024.
\newblock {EMMA}-500: Enhancing massively multilingual adaptation of large language models.
\newblock \emph{arXiv preprint arXiv:2409.17892}.

\bibitem[{Joshi et~al.(2025)Joshi, Dabre, Kanojia, Li, Zhan, Haffari, and Dippold}]{Aditya2025NLP4dialects}
Aditya Joshi, Raj Dabre, Diptesh Kanojia, Zhuang Li, Haolan Zhan, Gholamreza Haffari, and Doris Dippold. 2025.
\newblock \href {https://doi.org/10.1145/3712060} {Natural language processing for dialects of a language: A survey}.
\newblock \emph{ACM Comput. Surv.}, 57(6).

\bibitem[{Kargaran et~al.(2024)Kargaran, Yvon, and Sch{\"u}tze}]{kargaran-etal-2024-glotscript}
Amir~Hossein Kargaran, Fran{\c{c}}ois Yvon, and Hinrich Sch{\"u}tze. 2024.
\newblock \href {https://aclanthology.org/2024.lrec-main.687/} {{G}lot{S}cript: A resource and tool for low resource writing system identification}.
\newblock In \emph{Proceedings of the 2024 Joint International Conference on Computational Linguistics, Language Resources and Evaluation (LREC-COLING 2024)}, pages 7774--7784, Torino, Italia. ELRA and ICCL.

\bibitem[{Kuzhuget et~al.(2024)Kuzhuget, Mongush, and Oorzhak}]{wmt24-tuvan}
Ali Kuzhuget, Airana Mongush, and Nachyn-Enkhedorzhu Oorzhak. 2024.
\newblock Enhancing {Tuvan} language resources through the {FLORES} dataset.
\newblock In \emph{Proceedings of the Ninth Conference on Machine Translation}, Miami, USA. Association for Computational Linguistics.

\bibitem[{Kwon et~al.(2023)Kwon, Li, Zhuang, Sheng, Zheng, Yu, Gonzalez, Zhang, and Stoica}]{kwon2023efficient}
Woosuk Kwon, Zhuohan Li, Siyuan Zhuang, Ying Sheng, Lianmin Zheng, Cody~Hao Yu, Joseph~E. Gonzalez, Hao Zhang, and Ion Stoica. 2023.
\newblock Efficient memory management for large language model serving with pagedattention.
\newblock In \emph{Proceedings of the ACM SIGOPS 29th Symposium on Operating Systems Principles}.

\bibitem[{Lison and Tiedemann(2016)}]{lison-tiedemann-2016-opensubtitles2016}
Pierre Lison and J{\"o}rg Tiedemann. 2016.
\newblock \href {https://aclanthology.org/L16-1147/} {{O}pen{S}ubtitles2016: Extracting large parallel corpora from movie and {TV} subtitles}.
\newblock In \emph{Proceedings of the Tenth International Conference on Language Resources and Evaluation ({LREC}`16)}, pages 923--929, Portoro{\v{z}}, Slovenia. European Language Resources Association (ELRA).

\bibitem[{Ma et~al.(2024)Ma, ImaniGooghari, Ye, Pei, Asgari, and Sch{\"u}tze}]{ma-etal-2024-taxi1500}
Chunlan Ma, Ayyoob ImaniGooghari, Haotian Ye, Renhao Pei, Ehsaneddin Asgari, and Hinrich Sch{\"u}tze. 2024.
\newblock {Taxi1500}: A multilingual dataset for text classification in 1500 languages.
\newblock \emph{arXiv preprint arXiv:2305.08487}.

\bibitem[{Mamasaidov and Shopulatov(2024)}]{wmt24-karakalpak}
Mukhammadsaid Mamasaidov and Abror Shopulatov. 2024.
\newblock {Open Language Data Initiative}: Advancing low-resource machine translation for {Karakalpak}.
\newblock In \emph{Proceedings of the Ninth Conference on Machine Translation}, Miami, USA. Association for Computational Linguistics.

\bibitem[{Mayer and Cysouw(2014)}]{mayer-cysouw-2014-creating}
Thomas Mayer and Michael Cysouw. 2014.
\newblock Creating a massively parallel {B}ible corpus.
\newblock In \emph{Proceedings of the Ninth International Conference on Language Resources and Evaluation ({LREC}'14)}, pages 3158--3163, Reykjavik, Iceland. European Language Resources Association.

\bibitem[{{NLLB Team} et~al.(2022){NLLB Team}, Costa-juss{\`a}, Cross, {\c{C}}elebi, Elbayad, Heafield, Heffernan, Kalbassi, Lam, Licht, Maillard, Sun, Wang, Wenzek, Youngblood, Akula, Barrault, Gonzalez, Hansanti, Hoffman, Jarrett, Sadagopan, Rowe, Spruit, Tran, Andrews, Ayan, Bhosale, Edunov, Fan, Gao, Goswami, Guzm{\'a}n, Koehn, Mourachko, Ropers, Saleem, Schwenk, and Wang}]{nllb2022}
{NLLB Team}, Marta~R. Costa-juss{\`a}, James Cross, Onur {\c{C}}elebi, Maha Elbayad, Kenneth Heafield, Kevin Heffernan, Elahe Kalbassi, Janice Lam, Daniel Licht, Jean Maillard, Anna Sun, Skyler Wang, Guillaume Wenzek, Al~Youngblood, Bapi Akula, Lo{\"\i}c Barrault, Gabriel~M. Gonzalez, Prangthip Hansanti, John Hoffman, Semarley Jarrett, Kaushik~R. Sadagopan, Dirk Rowe, Shannon Spruit, Chau Tran, Pierre Andrews, Necip~Fazil Ayan, Shruti Bhosale, Sergey Edunov, Angela Fan, Cynthia Gao, Vedanuj Goswami, Francisco Guzm{\'a}n, Philipp Koehn, Alexandre Mourachko, Christophe Ropers, Safiyyah Saleem, Holger Schwenk, and Jeff Wang. 2022.
\newblock No language left behind: Scaling human-centered machine translation.
\newblock \emph{arXiv preprint arXiv:2207.04672}.

\bibitem[{{NLLB Team} et~al.(2024){NLLB Team}, Costa-juss{\`a}, Cross, {\c{C}}elebi, Elbayad, Heafield, Heffernan, Kalbassi, Lam, Licht, Maillard, Sun, Wang, Wenzek, Youngblood, Akula, Barrault, Gonzalez, Hansanti, Hoffman, Jarrett, Sadagopan, Rowe, Spruit, Tran, Andrews, Ayan, Bhosale, Edunov, Fan, Gao, Goswami, Guzm{\'a}n, Koehn, Mourachko, Ropers, Saleem, Schwenk, and Wang}]{nllb-24}
{NLLB Team}, Marta~R. Costa-juss{\`a}, James Cross, Onur {\c{C}}elebi, Maha Elbayad, Kenneth Heafield, Kevin Heffernan, Elahe Kalbassi, Janice Lam, Daniel Licht, Jean Maillard, Anna Sun, Skyler Wang, Guillaume Wenzek, Al~Youngblood, Bapi Akula, Loic Barrault, Gabriel~Mejia Gonzalez, Prangthip Hansanti, John Hoffman, Semarley Jarrett, Kaushik~Ram Sadagopan, Dirk Rowe, Shannon Spruit, Chau Tran, Pierre Andrews, Necip~Fazil Ayan, Shruti Bhosale, Sergey Edunov, Angela Fan, Cynthia Gao, Vedanuj Goswami, Francisco Guzm{\'a}n, Philipp Koehn, Alexandre Mourachko, Christophe Ropers, Safiyyah Saleem, Holger Schwenk, and Jeff Wang. 2024.
\newblock \href {https://doi.org/10.1038/s41586-024-07335-x} {Scaling neural machine translation to 200 languages}.
\newblock \emph{Nature}, 630(8018):841--846.

\bibitem[{OpenAI(2023)}]{openai2023gpt4}
OpenAI. 2023.
\newblock Gpt-4 technical report.
\newblock \emph{arXiv preprint arXiv:2303.08774}.

\bibitem[{Pan et~al.(2017)Pan, Zhang, May, Nothman, Knight, and Ji}]{pan-etal-2017-cross}
Xiaoman Pan, Boliang Zhang, Jonathan May, Joel Nothman, Kevin Knight, and Heng Ji. 2017.
\newblock \href {https://doi.org/10.18653/v1/P17-1178} {Cross-lingual name tagging and linking for 282 languages}.
\newblock In \emph{Proceedings of the 55th Annual Meeting of the Association for Computational Linguistics (Volume 1: Long Papers)}, pages 1946--1958, Vancouver, Canada. Association for Computational Linguistics.

\bibitem[{Perez-Ortiz et~al.(2024)Perez-Ortiz, S{\'a}nchez-Martínez, S{\'a}nchez-Cartagena, Esplà-Gomis, Jimenez, Oliver, Aventín-Boya, Pardos, Valdés, Socasau, and Martínez}]{wmt24-spain}
Juan~Antonio Perez-Ortiz, Felipe S{\'a}nchez-Martínez, Víctor~M. S{\'a}nchez-Cartagena, Miquel Esplà-Gomis, Aaron~Galiano Jimenez, Antoni Oliver, Claudi Aventín-Boya, Alejandro Pardos, Cristina Valdés, Jus{\'e}p Loís~Sans Socasau, and Juan~Pablo Martínez. 2024.
\newblock Expanding the flores+ multilingual benchmark with translations for {Aragonese, Aranese, Asturian, and Valencian}.
\newblock In \emph{Proceedings of the Ninth Conference on Machine Translation}, Miami, USA. Association for Computational Linguistics.

\bibitem[{Poelman and de~Lhoneux(2024)}]{poelman2024rolesenglishevaluatingmultilingual}
Wessel Poelman and Miryam de~Lhoneux. 2024.
\newblock \href {https://arxiv.org/abs/2412.08392} {The roles of english in evaluating multilingual language models}.
\newblock \emph{Preprint}, arXiv:2412.08392.

\bibitem[{Post(2018)}]{post2018call}
Matt Post. 2018.
\newblock A call for clarity in reporting bleu scores.
\newblock \emph{arXiv preprint arXiv:1804.08771}.

\bibitem[{Singh et~al.(2024{\natexlab{a}})Singh, Romanou, Fourrier, Adelani, and \textit{et al.}}]{singh-etal-2024-globalmmlu}
Shivalika Singh, Angelika Romanou, Cl{\'e}mentine Fourrier, David~Ifeoluwa Adelani, and \textit{et al.} 2024{\natexlab{a}}.
\newblock Global {MMLU}: Understanding and addressing cultural and linguistic biases in multilingual evaluation.
\newblock \emph{arXiv preprint arXiv:2412.03304}.

\bibitem[{Singh et~al.(2024{\natexlab{b}})Singh, Vargas, D'souza, Karlsson, Mahendiran, Ko, Shandilya, Patel, Mataciunas, O'Mahony, Zhang, Hettiarachchi, Wilson, Machado, Moura, Krzemi{\'n}ski, Fadaei, Ergun, Okoh, Alaagib, Mudannayake, Alyafeai, Chien, Ruder, Guthikonda, Alghamdi, Gehrmann, Muennighoff, Bartolo, Kreutzer, {\"U}st{\"u}n, Fadaee, and Hooker}]{singh-etal-2024-aya}
Shivalika Singh, Freddie Vargas, Daniel D'souza, B{\"o}rje Karlsson, Abinaya Mahendiran, Wei-Yin Ko, Herumb Shandilya, Jay Patel, Deividas Mataciunas, Laura O'Mahony, Mike Zhang, Ramith Hettiarachchi, Joseph Wilson, Marina Machado, Luisa Moura, Dominik Krzemi{\'n}ski, Hakimeh Fadaei, Irem Ergun, Ifeoma Okoh, Aisha Alaagib, Oshan Mudannayake, Zaid Alyafeai, Vu~Chien, Sebastian Ruder, Surya Guthikonda, Emad Alghamdi, Sebastian Gehrmann, Niklas Muennighoff, Max Bartolo, Julia Kreutzer, Ahmet {\"U}st{\"u}n, Marzieh Fadaee, and Sara Hooker. 2024{\natexlab{b}}.
\newblock Aya dataset: An open-access collection for multilingual instruction tuning.
\newblock In \emph{Proceedings of the 62nd Annual Meeting of the Association for Computational Linguistics (Volume 1: Long Papers)}, pages 11521--11567, Bangkok, Thailand. Association for Computational Linguistics.

\bibitem[{Tan et~al.(2024)Tan, Hansanti, Wood, Yu, Ropers, and Costa-jussà}]{tan2024massivemultilingualholisticbias}
Xiaoqing~Ellen Tan, Prangthip Hansanti, Carleigh Wood, Bokai Yu, Christophe Ropers, and Marta~R. Costa-jussà. 2024.
\newblock \href {https://arxiv.org/abs/2407.00486} {Towards massive multilingual holistic bias}.
\newblock \emph{Preprint}, arXiv:2407.00486.

\bibitem[{Tiedemann(2020)}]{tiedemann-2020-tatoeba}
J{\"o}rg Tiedemann. 2020.
\newblock The tatoeba translation challenge -- realistic data sets for low resource and multilingual {MT}.
\newblock In \emph{Proceedings of the Fifth Conference on Machine Translation}, pages 1174--1182, Online. Association for Computational Linguistics.

\bibitem[{Touvron et~al.(2023)Touvron, Martin, Stone, Albert, Almahairi, Babaei, Bashlykov, Batra, Bhargava, Bhosale, Bikel, Blecher, Ferrer, Chen, Cucurull, Esiobu, Fernandes, Fu, Fu, Fuller, Gao, Goswami, Goyal, Hartshorn, Hosseini, Hou, Inan, Kardas, Kerkez, Khabsa, Kloumann, Korenev, Koura, Lachaux, Lavril, Lee, Liskovich, Lu, Mao, Martinet, Mihaylov, Mishra, Molybog, Nie, Poulton, Reizenstein, Rungta, Saladi, Schelten, Silva, Smith, Subramanian, Tan, Tang, Taylor, Williams, Kuan, Xu, Yan, Zarov, Zhang, Fan, Kambadur, Narang, Rodriguez, Stojnic, Edunov, and Scialom}]{llama2}
Hugo Touvron, Louis Martin, Kevin Stone, Peter Albert, Amjad Almahairi, Yasmine Babaei, Nikolay Bashlykov, Soumya Batra, Prajjwal Bhargava, Shruti Bhosale, Dan Bikel, Lukas Blecher, Cristian~Canton Ferrer, Moya Chen, Guillem Cucurull, David Esiobu, Jude Fernandes, Jeremy Fu, Wenyin Fu, Brian Fuller, Cynthia Gao, Vedanuj Goswami, Naman Goyal, Anthony Hartshorn, Saghar Hosseini, Rui Hou, Hakan Inan, Marcin Kardas, Viktor Kerkez, Madian Khabsa, Isabel Kloumann, Artem Korenev, Punit~Singh Koura, Marie-Anne Lachaux, Thibaut Lavril, Jenya Lee, Diana Liskovich, Yinghai Lu, Yuning Mao, Xavier Martinet, Todor Mihaylov, Pushkar Mishra, Igor Molybog, Yixin Nie, Andrew Poulton, Jeremy Reizenstein, Rashi Rungta, Kalyan Saladi, Alan Schelten, Ruan Silva, Eric~Michael Smith, Ranjan Subramanian, Xiaoqing~Ellen Tan, Binh Tang, Ross Taylor, Adina Williams, Jian~Xiang Kuan, Puxin Xu, Zheng Yan, Iliyan Zarov, Yuchen Zhang, Angela Fan, Melanie Kambadur, Sharan Narang, Aurelien Rodriguez, Robert Stojnic, Sergey Edunov, and Thomas
  Scialom. 2023.
\newblock \href {https://arxiv.org/abs/2307.09288} {Llama 2: Open foundation and fine-tuned chat models}.
\newblock \emph{Preprint}, arXiv:2307.09288.

\bibitem[{Varab and Schluter(2021)}]{varab-schluter-2021-massivesumm}
Daniel Varab and Natalie Schluter. 2021.
\newblock \href {https://doi.org/10.18653/v1/2021.emnlp-main.797} {{M}assive{S}umm: a very large-scale, very multilingual, news summarisation dataset}.
\newblock In \emph{Proceedings of the 2021 Conference on Empirical Methods in Natural Language Processing}, pages 10150--10161, Online and Punta Cana, Dominican Republic. Association for Computational Linguistics.

\bibitem[{Wolf et~al.(2020)Wolf, Debut, Sanh, Chaumond, Delangue, Moi, Cistac, Rault, Louf, Funtowicz, Davison, Shleifer, von Platen, Ma, Jernite, Plu, Xu, Le~Scao, Gugger, Drame, Lhoest, and Rush}]{wolf-etal-2020-transformers}
Thomas Wolf, Lysandre Debut, Victor Sanh, Julien Chaumond, Clement Delangue, Anthony Moi, Pierric Cistac, Tim Rault, Remi Louf, Morgan Funtowicz, Joe Davison, Sam Shleifer, Patrick von Platen, Clara Ma, Yacine Jernite, Julien Plu, Canwen Xu, Teven Le~Scao, Sylvain Gugger, Mariama Drame, Quentin Lhoest, and Alexander Rush. 2020.
\newblock \href {https://doi.org/10.18653/v1/2020.emnlp-demos.6} {Transformers: State-of-the-art natural language processing}.
\newblock In \emph{Proceedings of the 2020 Conference on Empirical Methods in Natural Language Processing: System Demonstrations}, pages 38--45, Online. Association for Computational Linguistics.

\bibitem[{Yang et~al.(2024)Yang, Yang, Hui, Zheng, Yu, Zhou, Li, Li, Liu, Huang, Dong, Wei, Lin, Tang, Wang, Yang, Tu, Zhang, Ma, Yang, Xu, Zhou, Bai, He, Lin, Dang, Lu, Chen, Yang, Li, Xue, Ni, Zhang, Wang, Peng, Men, Gao, Lin, Wang, Bai, Tan, Zhu, Li, Liu, Ge, Deng, Zhou, Ren, Zhang, Wei, Ren, Liu, Fan, Yao, Zhang, Wan, Chu, Liu, Cui, Zhang, Guo, and Fan}]{qwen2}
An~Yang, Baosong Yang, Binyuan Hui, Bo~Zheng, Bowen Yu, Chang Zhou, Chengpeng Li, Chengyuan Li, Dayiheng Liu, Fei Huang, Guanting Dong, Haoran Wei, Huan Lin, Jialong Tang, Jialin Wang, Jian Yang, Jianhong Tu, Jianwei Zhang, Jianxin Ma, Jianxin Yang, Jin Xu, Jingren Zhou, Jinze Bai, Jinzheng He, Junyang Lin, Kai Dang, Keming Lu, Keqin Chen, Kexin Yang, Mei Li, Mingfeng Xue, Na~Ni, Pei Zhang, Peng Wang, Ru~Peng, Rui Men, Ruize Gao, Runji Lin, Shijie Wang, Shuai Bai, Sinan Tan, Tianhang Zhu, Tianhao Li, Tianyu Liu, Wenbin Ge, Xiaodong Deng, Xiaohuan Zhou, Xingzhang Ren, Xinyu Zhang, Xipin Wei, Xuancheng Ren, Xuejing Liu, Yang Fan, Yang Yao, Yichang Zhang, Yu~Wan, Yunfei Chu, Yuqiong Liu, Zeyu Cui, Zhenru Zhang, Zhifang Guo, and Zhihao Fan. 2024.
\newblock \href {https://arxiv.org/abs/2407.10671} {Qwen2 technical report}.
\newblock \emph{Preprint}, arXiv:2407.10671.

\bibitem[{Yu et~al.(2024)Yu, Shi, Zhou, and Haberland}]{wmt24-wu}
Hongjian Yu, Yiming Shi, Zherui Zhou, and Christopher Haberland. 2024.
\newblock Machine translation evaluation benchmark for {Wu}.
\newblock In \emph{Proceedings of the Ninth Conference on Machine Translation}, Miami, USA. Association for Computational Linguistics.

\bibitem[{Zhang et~al.(2024)Zhang, Gao, Zhu, Chen, Huang, Han, Feng, Deng, and Huang}]{zhang2024getting}
Shimao Zhang, Changjiang Gao, Wenhao Zhu, Jiajun Chen, Xin Huang, Xue Han, Junlan Feng, Chao Deng, and Shujian Huang. 2024.
\newblock Getting more from less: Large language models are good spontaneous multilingual learners.
\newblock In \emph{Proceedings of the 2024 Conference on Empirical Methods in Natural Language Processing (EMNLP)}, pages 8037--8051. Association for Computational Linguistics.

\end{thebibliography}

\clearpage %
\appendix

\section{Benchmark Settings}
\label{app:benchmarks}

\subsection{Intrinsic Evaluation}

Given the input $X = (x_0, x_1, \ldots, x_{n_t})$, the negative log-likelihood (NLL) is defined as:

\begin{equation}
    \text{NLL} = -\sum_{i=1}^{n_t} \log p_{\theta}(x_i | x_{<i})
\end{equation}
while perplexity (PPL) is computed as:
\begin{equation}
    \text{PPL} = \exp\{-\frac{1}{n_t}\sum_{i=1}^{n_t} \log p_{\theta}(x_i | x_{<i})\}
\end{equation}

Intuitively, PPL evaluates a model’s ability to predict tokens in a given corpus, with lower values indicating better performance. In contrast, NLL measures the overall likelihood of the corpus under the model. Notably, due to its length normalization, PPL is directly influenced by the tokenization scheme, whereas NLL remains unaffected. Therefore, we use NLL for model comparisons to ensure consistency across models with different tokenization methods.

We compute NLL by concatenating the input sentences and applying a strided sliding window of size 1024.

\subsection{Machine Translation}

\paragraph{FLORES+} This work builds upon previous efforts on multilingual machine translation and evaluation datasets \cite{nllb-24,flores101-22,flores1-19,mt4nko-23,indictrans2-23,wmt24-spain,wmt24-4african,wmt24-emakhuwa,wmt24-tuvan,wmt24-wu,wmt24-karakalpak,wmt24-erzya}.

\paragraph{AmericasNLP} Only the development set is used, as the test set is not disclosed. Note that this dataset is aligned with Spanish, but not English.
\paragraph{Tatoeba (v2023-09-26)} We keep only test sets with over 1{,}000 sentences.

\paragraph{BLEU} In our experiments, BLEU scores are computed via \texttt{SacreBLEU}~\citep{post2018call} with the \texttt{flores200} tokenizer to quantify translation quality. The BLEU signature is:

{\small\ttfamily
nrefs:1 | case:mixed | eff:no | tok:flores200 | smooth:exp | version:2.4.2
}

\paragraph{COMET} Users can specify the customized model in the configuration file. The default model is \href{https://huggingface.co/Unbabel/wmt22-comet-da}{\texttt{Unbabel/wmt22-comet-da}}.

\paragraph{ChrF with Gender}  ChrF with gender is an evaluation metric that calculates the standard chrF score separately for sentences marked with different grammatical genders (masculine and feminine). By comparing these scores, one can assess whether a translation system favors one gender form over the other, thereby revealing potential gender bias in its outputs.

\subsection{Text Classification}

In classification tasks, the model predicts by ranking logits for each category; candidate labels are tokenized, and the label corresponding to the token with the highest probability is selected.

\section{Throughput Statistics}

\label{app:efficiency}

GlotEval provides a uniform pipeline for measuring both decoding-heavy and classification-style tasks across different languages, scripts, and hardware setups. According to efficiency analysis conducted in section~\ref{sec:efficiency}, table~\ref{tab:throughput-table-Nvidia} and~\ref{tab:throughput-table-AMD} show throughput results on both NVIDIA A100 40GB and AMD MI250X 64GB GPU environments. 

\begin{table*}[ht]
\centering
\resizebox{0.9\textwidth}{!}{%
\begin{tabular}{lcccccc}
\toprule
\multirow{2}{*}{\textbf{Language}} & \textbf{FLORES-200(Eng-X)} & \textbf{Aya} & \textbf{XLSum} & \textbf{SIB-200} & \textbf{Global-MMLU} & \textbf{WikiANN} \\
& (3-shot) & (0-shot) & (0-shot) & (3-shot) & (0-shot) & (3-shot) \\
\midrule
\textbf{French} (\texttt{fra\_Latn})& 854 / 0.88 = 969.55& 447 / 0.77 = 583.55 & 67 / 0.09 = 720.32 & 10 / 0.53 = 18.88 & 10 / 0.27 = 36.17& 70 / 2.36 = 29.60 \\
\textbf{Swahili} (\texttt{swa\_Latn})& 1174 / 0.92 = 1274.74& 812 / 0.80 = 1020.78 & 150 / 0.56 = 268.32 & 10 / 0.56 = 17.71 & 10 / 0.31 = 32.65& 61 / 1.97 = 30.91 \\
\textbf{Vietnamese} (\texttt{vie\_Latn})& 1206 / 0.92 = 1304.01& 443 / 0.76 = 581.40 & 172 / 0.74 = 233.62 & 10 / 0.48 = 20.99 & 10 / 0.26 = 37.87& 74 / 2.41 = 30.66 \\
\textbf{Indonesian} (\texttt{ind\_Latn})& 776 / 0.87 = 893.11& 259 / 0.75 = 346.56 & 308 / 0.75 = 411.16 & 10 / 0.53 = 18.91 & 10 / 0.28 = 35.65& 54 / 2.02 = 26.79 \\
\midrule
\textbf{Latin Scri.}& 4010 / 3.59 = 1116.99 & 1961 / 3.08 = 636.69 & 697 / 2.14 = 325.70 & 40 / 2.10 = 19.05 & 40 / 1.12 = 35.71 & 259 / 8.76 = 29.57 \\
\midrule
\textbf{Kyrgyz} (\texttt{kir\_Cyrl})& 1174 / 0.93 = 1259.10& 436 / 0.76 = 573.19 & 324 / 0.75 = 429.72 & 10 / 0.72 = 13.95 & 10 / 0.35 = 28.98& 72 / 4.20 = 17.16 \\
\textbf{Russian}(\texttt{rus\_Cyrl})& 1280 / 1.86 = 688.45& 551 / 0.77 = 712.23 & 339 / 0.67 = 507.16 & 10 / 0.53 = 18.92 & 10 / 0.28 = 35.25& 71 / 3.51 = 20.20 \\
\textbf{Serbian} (\texttt{srp\_Cyrl})& 1118 / 0.92 = 1207.56& 475 / 0.76 = 621.45 & 342 / 0.76 = 452.07 & 10 / 0.62 = 16.25 & 10 / 0.30 = 33.14& 48 / 1.94 = 24.76 \\
\textbf{Ukrainian} (\texttt{ukr\_Cyrl})& 1083 / 0.91 = 1191.05& 404 / 0.76 = 532.91 & 43 / 0.09 = 470.72 & 10 / 0.68 = 14.65 & 10 / 0.31 = 31.68& 132 / 7.43 = 17.78 \\
\midrule
\textbf{Cyrillic Scri.}& 4655 / 4.62 = 1007.58 & 1866 / 3.05 = 611.80 & 1048 / 2.27 = 461.67 & 40 / 2.55 = 15.69 & 40 / 1.24 = 32.26 & 323 / 17.08 = 18.91 \\
\midrule
\textbf{Arabic} (\texttt{arb\_Arab})& 852 / 0.87 = 974.46& 74 / 0.41 = 181.59 & 228 / 1.62 = 140.36 & 10 / 0.53 = 18.85 & 10 / 0.28 = 36.32& 76 / 2.75 = 27.65 \\
\textbf{Persian} (\texttt{fas\_Arab})& 852 / 0.89 = 958.99& 264 / 0.75 = 353.62 & 333 / 0.76 = 440.70 & 10 / 0.68 = 14.63 & 10 / 0.31 = 31.74& 54 / 12.48 = 4.32\\
\midrule
\textbf{Arabic Scri.}& 1704 / 1.76 = 968.18 & 338 / 1.16 = 291.38 & 561 / 2.38 = 235.71 & 20 / 1.21 = 16.53 & 20 / 0.59 = 33.90 & 130 / 15.23 = 8.54 \\
\midrule
\textbf{Bengali} (\texttt{ben\_Beng})& 1143 / 0.96 = 1190.74& 973 / 0.81 = 1195.97 & 260 / 0.71 = 366.53 & 10 / 1.26 = 7.91 & 10 / 0.45 = 21.99& 39 / 2.13 = 18.32 \\
\textbf{Hindi} (\texttt{hin\_Deva})& 1167 / 0.96 = 1210.17& 960 / 0.81 = 1182.68 & 223 / 0.75 = 296.00 & 10 / 1.10 = 9.07 & 10 / 0.39 = 25.62& 52 / 2.50 = 20.79 \\
\textbf{Nepali} (\texttt{npi\_Deva})& 1250 / 1.01 = 1247.45& 803 / 0.80 = 1009.63 & 231 / 0.60 = 384.25 & 10 / 1.02 = 9.78 & 10 / 0.41 = 24.57& 69 / 3.98 = 17.32\\
\midrule
\textbf{Devanagari}& 3560 / 2.93 = 1215.02 & 2736 / 2.42 = 1130.58 & 714 / 2.06 = 346.60 & 30 / 3.38 = 8.88 & 30 / 1.25 = 24.00 & 160 / 8.61 = 18.58 \\
\midrule
\textbf{Sinhala} (\texttt{sin\_Sinh})& 1280 / 1.04 = 1226.21& 1280 / 0.86 = 1485.77 & 103 / 0.17 = 601.67 & 10 / 1.57 = 6.38 & 10 / 0.52 =19.38& 69 / 5.43 = 12.70 \\
\midrule
\textbf{Telugu} (\texttt{tel\_Telu})& 1208 / 1.02 = 1188.70& 559 / 0.80 = 697.54 & 74 / 0.14 = 537.25 & 10 / 1.57 = 6.38 & 10 / 0.55 = 18.21& 71 / 8.01 = 8.86 \\
\midrule
\textbf{Amharic} (\texttt{amh\_Ethi})& 1280 / 1.00 = 1278.40& 1280 / 0.85 = 1498.47 & 65 / 0.09 = 700.75 & 10 / 1.00 = 9.95 & 10 / 7.37 = 1.36& 53 / 10.31 = 5.14 \\
\midrule
\textbf{Japanese} (\texttt{jpn\_Jpan})& 714 / 0.87 = 820.25& 152 / 0.21 = 707.20 & 274 / 0.75 = 365.11 & 10 / 0.48 = 21.01 & 10 / 0.28 = 35.99& 389 / 33.70 = 11.54 \\
\midrule
\textbf{Korean} (\texttt{kor\_Hang})& 1016 / 0.90 = 1129.38& 284 / 0.76 = 374.84 & 59 / 0.12 = 493.29 & 10 / 0.54 = 18.58 & 10 / 0.27 = 36.78& 91 / 5.20 = 17.50 \\
\midrule
\textbf{Chinese} (\texttt{zho\_Hans})& 676 / 0.87 = 780.69& 403 / 0.62 = 651.94 & 59 / 0.12 = 491.30 & 10 / 0.41 = 24.11 & 10 / 0.26 =37.60& 419 / 42.26 = 9.91 \\
\bottomrule
\end{tabular}
}%
\caption{Throughput with NVIDIA A100 40GB GPU.
Each cell contains: \emph{\(\frac{\#\text{generated tokens}}{\text{wall time (seconds)}} = \text{average tokens/s}\)}.}
\label{tab:throughput-table-Nvidia}
\end{table*}
\begin{table*}[ht]
\centering
\resizebox{0.9\textwidth}{!}{%
\begin{tabular}{lcccccc}
\toprule
\multirow{2}{*}{\textbf{Language}} & \textbf{FLORES-200(Eng-X)} & \textbf{Aya} & \textbf{XLSum} & \textbf{SIB-200} & \textbf{Global-MMLU} & \textbf{WikiANN} \\
& (3-shot) & (0-shot) & (0-shot) & (3-shot) & (0-shot) & (3-shot) \\
\midrule
\textbf{French} (\texttt{fra\_Latn})& 800 / 1.53 = 524.33 & 409 / 1.34 = 304.24 & 164 / 1.01 = 161.69 & 10 / 29.00 = 0.34 & 10 / 39.30 = 0.25 & 70 / 38.18 = 1.83 \\
\textbf{Swahili} (\texttt{swa\_Latn})& 1039 / 1.55 = 670.79 & 136 / 0.43 = 317.94 & 226 / 0.93 = 244.26 & 10 / 26.71 = 0.37 & 10 / 38.61 = 0.26 & 61 / 38.21 = 1.60 \\
\textbf{Vietnamese} (\texttt{vie\_Latn})& 932 / 1.53 = 608.26 & 675 / 1.39 = 485.18 & 58 / 0.15 = 379.43 & 10 / 31.58 = 0.32 & 10 / 39.46 = 0.25 & 74 / 38.13 = 1.94 \\
\textbf{Indonesian} (\texttt{ind\_Latn})& 1076 / 1.52 = 706.44 & 779 / 1.40 = 555.64 & 262 / 1.29 = 203.48 & 10 / 29.33 = 0.34 & 10 / 39.14 = 0.26 & 54 / 37.16 = 1.45 \\
\midrule
\textbf{Latin Scri.}& 3847 / 6.13 = 627.57 & 1999 / 4.56 = 438.38 & 710 / 3.38 = 210.06 & 40 / 29.20 = 1.37 & 40 / 39.22 = 1.02 & 259 / 37.98 = 6.82 \\
\midrule
\textbf{Kyrgyz} (\texttt{kir\_Cyrl})& 1051 / 1.57 = 669.63 & 344 / 1.32 = 261.10 & 444 / 1.36 = 325.96 & 10 / 17.18 = 0.58 & 10 / 37.68 = 0.27 & 72 / 23.48 = 3.07 \\
\textbf{Russian}(\texttt{rus\_Cyrl})& 1280 / 1.86 = 686.47 & 442 / 1.37 = 322.04 & 243 / 1.03 = 234.98 & 10 / 29.37 = 0.34 & 10 / 39.04 = 0.26 & 71 / 30.55 = 2.32 \\
\textbf{Serbian} (\texttt{srp\_Cyrl})& 1210 / 1.58 = 767.56 & 560 / 1.38 = 406.04 & 261 / 1.17 = 222.39 & 10 / 19.57 = 0.51 & 10 / 38.61 = 0.26 & 48 / 36.64 = 1.31 \\
\textbf{Ukrainian} (\texttt{ukr\_Cyrl})& 939 / 1.55 = 607.67 & 378 / 1.33 = 284.88 & 103 / 0.42 = 244.48 & 10 / 17.34 = 0.58 & 10 / 38.31 = 0.26 & 132 / 23.54 = 5.61 \\
\midrule
\textbf{Cyrillic Scri.}& 4480 / 6.56 = 682.93 & 1724 / 5.40 = 319.26 & 1051 / 3.98 = 264.07 & 40 / 19.90 = 2.01 & 40 / 38.10 = 1.05 & 323 / 26.24 = 12.31 \\
\midrule
\textbf{Arabic} (\texttt{arb\_Arab})& 919 / 1.54 = 595.36 & 160 / 1.24 = 129.25 & 83 / 0.26 = 318.11 & 10 / 29.06 = 0.34 & 10 / 39.23 = 0.25 & 76 / 37.56 = 2.02 \\
\textbf{Persian} (\texttt{fas\_Arab})& 929 / 1.55 = 600.20 & 16 / 0.12 = 131.16 & 184 / 1.21 = 152.61 & 10 / 17.43 = 0.57 & 10 / 38.25 = 0.26 & 54 / 13.24 = 4.07\\
\midrule
\textbf{Arabic Scri.}& 1848 / 3.09 = 598.06 & 176 / 1.36 = 129.41 & 267 / 1.47 = 181.63 & 20 / 21.98 = 0.91 & 20 / 39.22 = 0.51 & 130 / 21.35 = 6.09 \\
\midrule
\textbf{Bengali} (\texttt{ben\_Beng})& 1130 / 1.62 = 698.59 & 1026 / 1.40 = 734.29 & 178 / 1.21 = 147.66 & 10 / 11.17 = 0.90 & 10 / 27.49 = 0.36 & 39 / 28.34 = 1.38 \\
\textbf{Hindi} (\texttt{hin\_Deva})& 1160 / 1.62 = 714.58 & 650 / 1.41 = 462.11 & 186 / 1.21 = 154.24 & 10 / 12.17 = 0.82 & 10 / 34.96 = 0.29 & 52 / 31.38 = 1.66 \\
\textbf{Nepali} (\texttt{npi\_Deva})& 1280 / 1.66 = 768.85 & 1126 / 1.40 = 805.59 & 275 / 1.10 = 250.46 & 10 / 13.00 = 0.77 & 10 / 34.68 = 0.29 & 69 / 2.77 = 24.87\\
\midrule
\textbf{Devanagari}& 3570 / 4.90 = 728.57 & 2802 / 4.21 = 665.56 & 639 / 3.52 = 181.53 & 30 / 12.05 = 2.49 & 30 / 31.91 = 0.94 & 160 / 5.73 = 27.91 \\
\midrule
\textbf{Sinhala} (\texttt{sin\_Sinh})& 1280 / 1.76 = 727.47 & 1223 / 1.42 = 858.97 & 140 / 0.93 = 151.08 & 10 / 9.15 = 1.09 & 10 / 25.60 = 0.39 & 69 / 15.70 = 4.40 \\
\midrule
\textbf{Telugu} (\texttt{tel\_Telu})& 1280 / 1.73 = 737.77 & 507 / 1.45 = 348.78 & 198 / 0.92 = 214.92 & 10 / 9.14 = 1.09 & 10 / 24.87 = 0.40 & 71 / 11.99 = 5.92 \\
\midrule
\textbf{Amharic} (\texttt{amh\_Ethi})& 1280 / 1.66 = 772.22 & 1153 / 1.40 = 821.85 & 211 / 1.12 = 189.18 & 10 / 13.04 = 0.77 & 10 / 34.30 = 0.29 & 53 / 32.18 = 1.65 \\
\midrule
\textbf{Japanese} (\texttt{jpn\_Jpan})& 690 / 1.51 = 458.09 & 266 / 1.27 = 209.63 & 250 / 1.02 = 244.87 & 10 / 31.85 = 0.31 & 10 / 39.23 = 0.25 & 389 / 14.98 = 25.96 \\
\midrule
\textbf{Korean} (\texttt{kor\_Hang})& 973 / 1.53 = 633.96 & 468 / 1.38 = 340.21 & 204 / 1.07 = 191.09 & 10 / 28.74 = 0.35 & 10 / 39.33 = 0.25 & 91 / 25.01 = 3.64 \\
\midrule
\textbf{Chinese} (\texttt{zho\_Hans})& 823 / 1.52 = 540.58 & 248 / 1.00 = 248.61 & 109 / 0.39 = 276.83 & 10 / 35.54 = 0.28 & 10 / 39.37 = 0.25 & 419 / 13.88 = 30.20 \\
\bottomrule
\end{tabular}
}%
\caption{Throughput with AMD MI250X 64GB GPU.
Each cell contains: \emph{\(\frac{\#\text{generated tokens}}{\text{wall time (seconds)}} = \text{average tokens/s}\)}.}
\label{tab:throughput-table-AMD}
\end{table*}

\end{document}